\documentclass{article} 
\usepackage{iclr2023_conference,times}

\usepackage{amsmath,amsfonts,bm}

\def\eqref#1{equation~\ref{#1}}

\def\1{\bm{1}}

\DeclareMathAlphabet{\mathsfit}{\encodingdefault}{\sfdefault}{m}{sl}
\SetMathAlphabet{\mathsfit}{bold}{\encodingdefault}{\sfdefault}{bx}{n}

\usepackage{hyperref}
\usepackage{url}
\usepackage{graphicx}

\title{Few-shot Diagnosis of Chest x-rays Using an Ensemble of Random Discriminative Subspaces}

\author{Kshitiz, Garvit Garg, Angshuman Paul\\
Department of Computer Science\\
Indian Institute of Technology Jodhpur\\
\texttt{\{kshitiz.1, garg.11, apaul\}@iitj.ac.in} \\
}

\iclrfinalcopy 
\begin{document}

\maketitle

\begin{abstract}
Due to the scarcity of annotated data in the medical domain, few-shot learning may be useful for medical image analysis tasks. We design a few-shot learning method using an ensemble of random subspaces for the diagnosis of chest x-rays (CXRs). Our design is computationally efficient and almost 1.8 times faster than method that uses the popular truncated singular value decomposition (t-SVD) for subspace decomposition. The proposed method is trained by minimizing a novel loss function that helps create well-separated clusters of training data in discriminative subspaces. As a result, minimizing the loss maximizes the distance between the subspaces, making them discriminative and assisting in better classification. Experiments on large-scale publicly available CXR datasets yield promising results. Code for the project will be available at \def\UrlFont{\bfseries\rmfamily} \url{https://github.com/Few-shot-Learning-on-chest-x-ray/fsl_subspace}.

\end{abstract}

\section{Introduction}

Few-shot Learning \cite{snell2017prototypical, chen2019closer, wang2020generalizing} aims to learn new tasks with small annotated samples per class. Most modern deep learning methods require a large number of annotated training samples. However, large annotated medical image datasets are often elusive because of various factors including cost and human efforts involved. Therefore, few-shot learning may be useful for medical image analysis. Few-shot learning techniques are often designed using either meta-learning \cite{finn2017model, zhang2018metagan, ren2018meta} or metric learning \cite{vinyals2016matching, simon2020adaptive} based approaches.

In this paper, we present a method for a few-shot diagnosis of chest x-ray (CXR) images using an ensemble of random subspaces. Our method consists of three stages, namely, a feature extraction module (FEM) that extracts visual feature vectors from x-ray images, a subspace embedding module (SEM) to project the feature vectors into multiple random discriminative subspaces, and a final decision module (FDM) that assigns a final class label to an input x-ray image based on the projections. The SEM in the proposed method aims to create an ensemble of discriminative subspaces to explore different combinations of visual features obtained from small data samples. To that end, our method generates subspaces having low similarity with each other. Such subspaces with low correlation are likely to aid ensemble learning \cite{breiman2001random}.

In this paper, we make the following major contributions: $i$) We propose a method for the few-shot chest x-ray diagnosis using an ensemble of random subspaces. $ii)$ Our method consists of multiple modules along with a novel loss component that makes these subspaces class discriminative, resulting in improved classification accuracy. $iii)$ The proposed method provides a faster alternative to existing computationally intensive subspace decomposition techniques such as truncated singular value decomposition (t-SVD) \cite{li2019tutorial}.

The rest of the paper is organized as follows: Section 2 provides the details of the proposed method. The experiments and results are described in Section 3. In Section 4, we conclude the paper. Finally, we describe the implementation details along with the ablation studies in the appendices.

\begin{figure*}[]
\centering
\scalebox{0.8}{
\includegraphics[width = \textwidth]{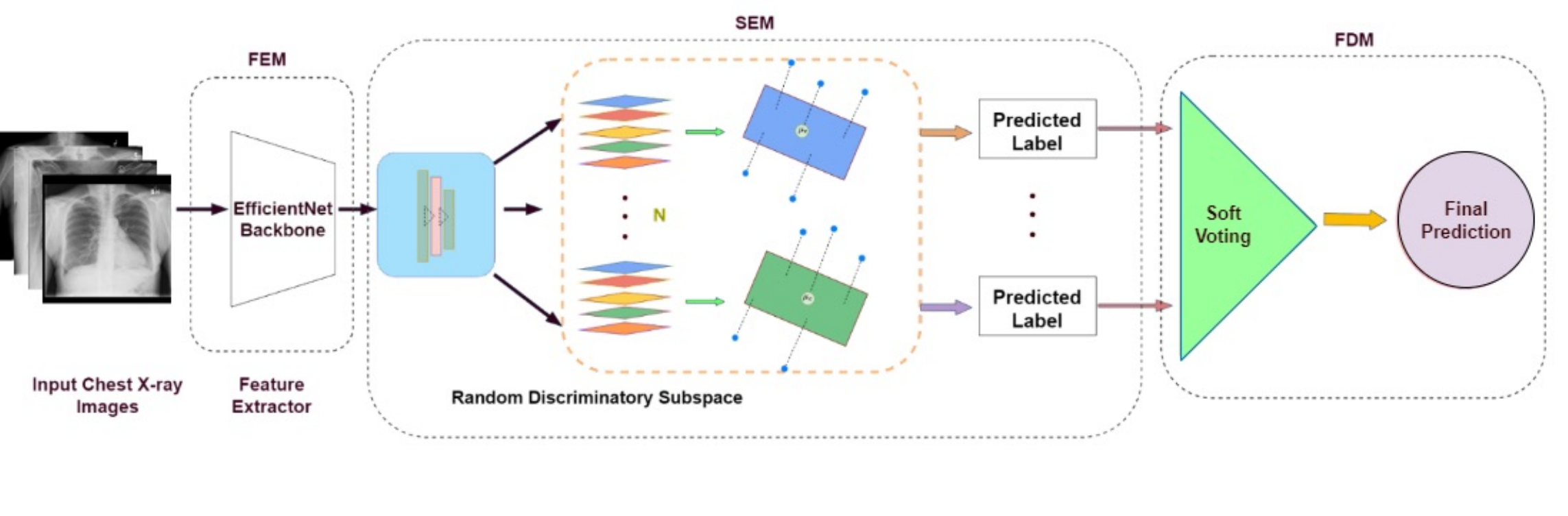}}
\vspace{-.1in}
\caption{Pipeline of the proposed method. Our model consists of a feature extraction module (FEM),  a subspace embedding module (SEM), and a final decision module (FDM). FEM extracts visual features from the chest x-ray images. SEM projects the extracted feature vectors into multiple random subspaces, each of which assigns a class label to the input x-ray. These labels are then used by FDM to produce the final output label.}
\label{fig1}
\end{figure*}

\section{Methods}

In the standard few-shot classification setting, we define a support set $S = (x_1, y_1),...,(x_K, y_K)$ and a query set $Q = (x'_1, y'_1),...,(x'_{K'}, y'_{K'})$ of $K$ and $K'$ labelled points respectively. Here each $x_i$ and $x'_i$ are $D-$dimensional feature vectors and $y_i$ and $y'_i$ are the corresponding class labels, respectively. The problem is formulated as $N$-way $K$-shot, where $N$ represents the number of classes in the support set, while $K$ denotes the number of examples of each class. The optimizations iterate in an episodic fashion taking one update at a time. In addition, the model parameters are updated by minimizing the loss function. Our model consists of three modules, namely the Feature extraction module (FEM), the Subspace embedding module (SEM), and the Final decision module (FDM). A block diagram of the proposed method is presented in Fig. \ref{fig1}.

\subsection{Feature extraction module (FEM)}

FEM is used for extracting visual features from chest x-rays. To design the FEM, we use a feature extractor, pre-trained on the ImageNet \cite{russakovsky2015imagenet} dataset. The FEM, designed using the EfficientNet-B7 backbone \cite{tan2019efficientnet} is further fine-tuned using x-ray images in concert with the other modules. The extracted feature vectors are then fed into the SEM.

\subsection{Subspace Embedding Module (SEM)}
\label{subsec:sem}
The subspace embedding module maps an input feature vector (obtained from FEM) to an ensemble of subspaces. Consider an embedding function $ E_{\theta, \nu}$ an ${D}$-dimensional input feature vector to a ${N}$-dimensional feature vector in subspace $\nu$. This embedding function is implemented through fully connected layers with parameters $\theta$. Consider class $j$. We calculate the mean embedding for class $j$ at subspace $\nu$ considering all the support points belonging to class $j$. The mean embedding is 
\begin{equation}
\lambda_{j}(\nu)=\frac{1}{C_j} \sum_{a_{i} \in j} E_{\theta, \nu}\left(a_{i}\right),
\end{equation}
where $a_i$ is the $i^{\textrm{th}}$ data point belonging to class $j$ and $ C_j $ is the number of support points in class $j$. The Euclidean distance between the mean embedding and the query point $q$ is

\begin{equation}
\label{eq:5}
R_j(q, \nu)=\left\|E_{\theta, \nu}\left(q\right)-\lambda_{{j}}(\nu)\right\|_2, 
\end{equation}
where, $\left\|.\right\|_2$ represent the L2-norm. Further, by applying the softmax on (\ref{eq:5}) in subspace $\nu$, we obtain the probability that query point $q$ belongs to class $j$ 

\begin{equation} 
\label{eq:3}
S_{j}(\nu)=\frac{e^{-R_{j}(\nu)}}{\sum_{j}e^{-R_{j}(\nu)}}.
\end{equation}
Thus, for the query point, we may find the class probabilities for all classes at each subspace. We aim to utilize complimentary information from the extracted feature vectors through an ensemble of subspaces. To that end, we want every pair of subspaces to be discriminative and have less similarities with each other. The similarities are captured through a discriminative loss which is minimized in our method. Consider subspaces $x$ and $y$. Let the parameters of the neural network layers mapping to these subspaces be $\theta_x$ and $\theta_y$, respectively. Then the discriminative loss is 
\begin{equation}
\label{eq:6}
\mathcal{L}_{dis}=\sum_{\forall x,y}\frac{\theta_x \cdot \theta_y}{\left|\theta_x\right|\left|\theta_y\right|},
\end{equation} 
where $\cdot$ indicates a scalar product. Minimization of the above loss function helps in making the subspaces discriminative. The class probabilities from ({\ref{eq:3}}) are next used by the final decision module. 

\subsection{Final decision module (FDM)}

The FDM uses the class probabilities from the subspaces. The class label with the highest vote from the subspaces is considered to be the final output label for an input CXR image. Using the output labels from FDM and the ground truth labels, we compute a cross-entropy loss $\mathcal{L}_{sup}$ for the support set images of the training classes. Also, using the output labels from FDM for the query images of the training classes, we compute a cross-entropy loss $\mathcal{L}_{qur}$. Hence, the total loss is   
\begin{equation} 
\label{eq:1}
\mathcal{L}=\mathcal{L}_{ {sup }}+\alpha \mathcal{L}_{ {qur }}+\beta \mathcal{L}_{ {dis}},
\end{equation} 
where $\alpha$ and $\beta$ are predefined constants. Our model is trained by minimizing the above loss.

\subsection{Testing}
We randomly select five image samples from each test class for testing and compute the cluster center based on those five images for each test class. Query image are then taken from the test classes, and our model projects each query images onto randomly selected subspaces. Every subspace assigns a class label to a query image as discussed in Section \ref{subsec:sem}. The final class label for a query image is obtained by soft voting using the class probabilities.

\begin{figure}[]
\centering
\scalebox{0.8}{
\includegraphics[width = \textwidth]{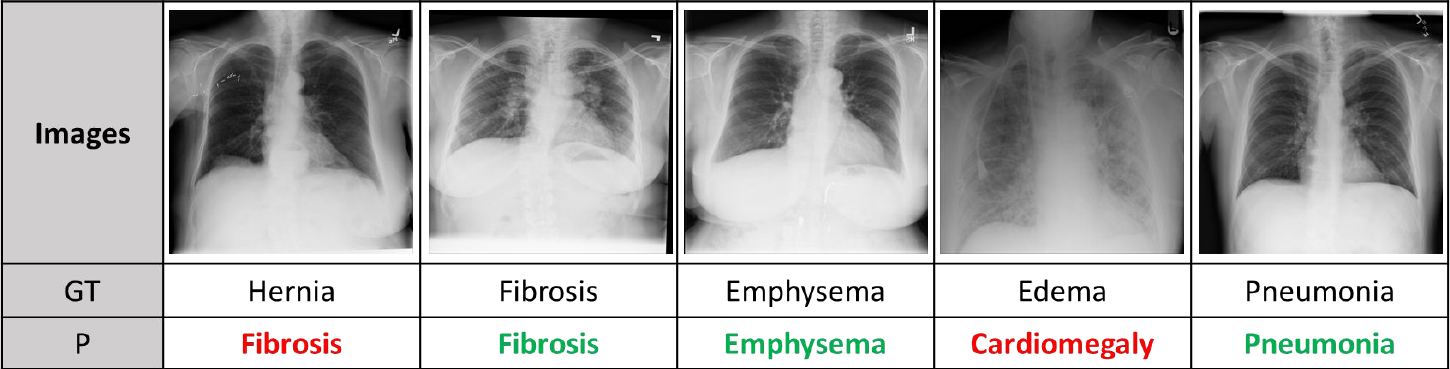}}
\caption{Sample results of the prediction (P) by the proposed method and the ground truth (GT) for images of novel classes. Correct and incorrect detections are marked in green and red, respectively.}
\label{figure3}
\end{figure}
\section{Results}

\subsection{Datasets and Experimental Settings}

We perform experiments on the publicly available NIH chest x-ray dataset \cite{wang2017chestx} which contains frontal CXR images with 14 different types of thoracic abnormalities. We only take images with single abnormalities for our experiments. Atelectasis, consolidation, infiltration, pneumothorax, fibrosis, effusion, pneumonia, pleural thickening, nodule, mass, hernia, edema, emphysema, and cardiomegaly are among the 14 thorax disorders or conditions studied. When a sample does not contain any abnormality, it is labeled as `NoFinding'. We have considered five groups with different combinations of training, validation, and test classes. At each group, three test classes, three validation classes, and nine training classes are selected randomly without replacement, ensuring that these classes do not overlap. We also ensure that the test classes across different groups do not overlap so that every class is used for testing in some experiment. A detailed description of the groups is presented in appendix \ref{sec:appendix_a} (see Table \ref{tab:Table3}). Furthermore, we evaluate our model's performance on multiple random subspaces by conducting experiments with varying numbers of subspaces under the same conditions. Our findings indicate that using 30 ($\nu$ = 30) subspaces leads to improved performance.

\begin{table}[h]
\scriptsize
\centering
\setlength{\tabcolsep}{2.5 pt} 
\renewcommand{\arraystretch}{1.25}
\caption{Accuracy of 3-way 5-shot classification with 95\% confidence interval on the different combinations of test/novel classes of the NIH dataset for different methods. The different combinations are represented by different groups mentioned in appendix \ref{sec:appendix_a}\\}
\scalebox{0.9}{
\resizebox{0.8\columnwidth}{!}{%
\begin{tabular}{|c|c|c|c|c|c|c|}
\hline
Group  & Abnormality        & ProtoNet   & MatchingNet & MAML       & DSN        & \textbf{Proposed} \\ \hline
       & Fibrosis           & 44.24±1.52 & 38.04±1.89  & 42.80±0.43 & 38.94±1.48 & 42.82±1.75        \\
Group1 & Hernia             & 44.34±1.50 & 38.24±1.89  & 16.85±0.30 & 43.38±1.51 & 30.10±1.62        \\
       & Pneumonia          & 45.68±1.55 & 30.43±1.89  & 44.89±0.44 & 50.28±1.57 & 40.88±1.66        \\ \hline
       & Mass               & 34.78±1.47 & 34.00±1.85  & 41.94±0.44 & 38.34±1.50 & 35.68±1.42        \\
Group2 & Nodule             & 34.86±1.46 & 34.60±1.87  & 14.65±0.28 & 35.84±1.48 & 35.96±1.48        \\
       & Pleural Thickening & 34.48±1.45 & 34.56±1.87  & 43.85±0.45 & 33.08±1.44 & 31.08±1.41        \\ \hline
       & Cardiomegaly       & 52.58±1.78 & 45.04±2.38  & 38.42±0.43 & 45.94±1.54 & 38.62±1.54        \\
Group3 & Edema              & 54.18±1.79 & 47.96±2.37  & 39.21±0.41 & 68.00±1.47 & 65.16±1.48        \\
       & Emphysema          & 54.64±1.77 & 47.04±2.41  & 34.60±0.43 & 49.08±1.60 & 42.04±1.61        \\ \hline
       & Consolidation      & 39.62±1.50 & 41.80±1.86  & 25.17±0.37 & 38.98±1.54 & 32.84±1.58        \\
Group4 & Effusion           & 41.26±1.60 & 40.06±1.84  & 22.30±0.35 & 42.82±1.54 & 38.20±1.61        \\
       & Pneumothorax       & 24.94±1.55 & 40.42±1.82  & 56.01±0.43 & 44.98±1.59 & 49.98±1.74        \\ \hline
       & Atelectasis        & 38.14±1.50 & 36.46±1.66  & 44.21±0.45 & 37.72±1.53 & 27.98±1.44        \\
Group5 & Infiltration       & 38.36±1.50 & 36.28±1.65  & 11.39±0.23 & 37.22±1.55 & 35.94±1.50        \\
       & No Finding         & 24.92±1.54 & 38.06±1.72  & 45.98±0.46 & 38.86±1.52 & 52.84±1.71        \\ \hline
\end{tabular}%
}}
\label{tab:Table1}
\end{table}

\begin{table}[h]
\centering
\tiny
\caption{Comparison of the training time (1 epoch) of the proposed method with the DSN \cite{simon2020adaptive} involving truncated SVD. The experiments have been executed using Nvidia-GeForce GTX 1080Ti GPU with 11,178 MiB memory. \\}
\label{tab:my-table}
\resizebox{0.8\columnwidth}{!}{
\begin{tabular}{cccc}
\hline
Method   & Approach                     & Train time    & Hardware specs \\ \hline
\textbf{Proposed} & Ensemble of Random Subspace  & \textbf{18 min 35 sec} & GTX 1080 Ti    \\ \hline
DSN      & Truncated Singular Value Decomposition (t-SVD) & 33 min 48 sec & GTX 1080 Ti    \\ \hline
\end{tabular}}
\label{tab:Table2}
\end{table}

\subsection{Comparative Performances}
We compare the performance of our method with several state-of-the-art few-shot learning methods, including Prototypical Network (ProtoNet) \cite{snell2017prototypical}, Matching Network (MatchingNet) \cite{vinyals2016matching}, MAML \cite{finn2017model}, and Adaptive Subspace (DSN) \cite{simon2020adaptive}. The results are evaluated using 15 query images for 3-way 5-shot classification in terms of mean average accuracy with a 95\% confidence interval and reported in Table \ref{tab:Table1} for the test/novel classes listed in the appendix \ref{sec:appendix_a} (see Table \ref{tab:Table3}). Notice that our approach yields better or comparable performance in terms of accuracy for a number of the novel classes. This shows the effectiveness of our method for the few-shot diagnosis of CXRs.

The proposed method is also faster than DSN which uses the t-SVD as the subspace decomposition technique. The running time of our method is compared in Table \ref{tab:Table2} with that of DSN \cite{simon2020adaptive}. It can be observed that the proposed method is approximately $\sim 1.8$ faster in a single epoch. This speed-up is due to the use of random subspaces in our method. A complexity analysis of the proposed method (see appendix \ref{sec:appendix_d}) with t-SVD demonstrates the advantage in terms of computational efficiency over t-SVD based methods.

\section{Conclusions}
We propose a few-shot learning model using random discriminative subspaces for chest x-ray diagnosis. The use of random subspaces makes our method a faster alternative to compute-intensive methods like t-SVD. Experiments and ablation studies demonstrate the utility of our method, which employs novel loss functions to aid in the generation of discriminative subspaces. In the future, we will explore the possibility of using auxiliary information about different abnormalities to aid the classification. We would also look into developing generalizable few-shot learning models for chest x-ray diagnosis.

\bibliographystyle{iclr2023_conference}
\bibliography{iclr2023_conference}

\pagebreak

\appendix
\section{Combination of abnormalities for different experiment}
\label{sec:appendix_a}

The description below details the combination of anomalies for different experiments, which are conducted separately for each group.

\begin{table*}[h]
\centering
\scriptsize
\setlength{\tabcolsep}{1.5pt} 
\renewcommand{\arraystretch}{1.25}
\caption{Combinations of the train/base, validation, and test/novel classes from Group1 to Group5 of the NIH dataset.\\}
\resizebox{0.9\columnwidth}{!}{
\begin{tabular}{|c|c|c|c|}
\hline
Group & Train/Base Class & Validation Class & Test/Novel Class \\ \hline
Group1 & \begin{tabular}[c]{@{}c@{}}Mass, Edema, Cardiomegaly,\\ Effusion, Infiltration, Nodule,\\ Emphysema, No Finding, Pneumothorax\end{tabular} & \begin{tabular}[c]{@{}c@{}}Atelectasis, Consolidation,\\ Pleural Thickening\end{tabular} & \begin{tabular}[c]{@{}c@{}}Hernia, Pneumonia, \\ Fibrosis\end{tabular} \\ \hline
Group2 & \begin{tabular}[c]{@{}c@{}}Effusion, Consolidation, Edema,\\ Cardiomegaly, No Finding, Atelectasis,\\ Infiltration, Emphysema, Pneumothorax\end{tabular} & \begin{tabular}[c]{@{}c@{}}Fibrosis, Hernia,\\ Pneumonia\end{tabular} & \begin{tabular}[c]{@{}c@{}}Mass, Nodule, \\ Pleural Thickening\end{tabular} \\ \hline
Group3 & \begin{tabular}[c]{@{}c@{}}Pneumothorax, Consolidation, Hernia,\\ No Finding, Atelectasis, Infiltration, \\ Effusion, Pneumonia, Fibrosis\end{tabular} & \begin{tabular}[c]{@{}c@{}}Mass, Nodule,\\ Pleural Thickening\end{tabular} & \begin{tabular}[c]{@{}c@{}}Emphysema, Edema, \\ Cardiomegaly\end{tabular} \\ \hline
Group4 & \begin{tabular}[c]{@{}c@{}}Infiltration, Hernia, Fibrosis, \\ No Finding, Atelectasis, Nodule, \\ Mass, Pneumonia, Pleural Thickening\end{tabular} & \begin{tabular}[c]{@{}c@{}}Emphysema, Edema, \\ Cardiomegaly\end{tabular} & \begin{tabular}[c]{@{}c@{}}Consolidation, Effusion,\\ Pneumothorax\end{tabular} \\ \hline
Group5 & \begin{tabular}[c]{@{}c@{}}Hernia, Fibrosis, Pneumonia, \\ Pleural Thickening, Nodule, Mass, \\ Emphysema, Edema, Cardiomegaly\end{tabular} & \begin{tabular}[c]{@{}c@{}}Consolidation, Effusion,\\ Pneumothorax\end{tabular} & \begin{tabular}[c]{@{}c@{}}Infiltration, Atelectasis,\\ No Finding\end{tabular} \\ \hline
\end{tabular}}
\label{tab:Table3}
\end{table*}

\section{Implementation Details}

The feature extraction module is designed using EfficientNet-B7 \cite{tan2019efficientnet}. The network module for projecting the input feature vector into subspaces comprises of linear and batch normalization layers. The 1000-dimensional feature space obtained from the input image using the feature extractor is then mapped to 512-dimension and further to 64-dimensional embedding. Using the chest x-rays as our training data, we perform end-to-end training on the ImageNet pre-trained model for 80 epochs consisting of 1000 batches with a single episode per batch with cross-entropy loss as the loss function. Further, we assume that the constants $\alpha$ and $\beta$ are equal to 1 for the loss calculation described in (\ref{eq:1}).

\section{Ablation Studies}

We perform several ablation studies to evaluate the effectiveness of different aspects of the model that contribute to yielding better performance over existing techniques. 

\subsection{On the Components of the Loss Functions}

We look into the effect of different components of the loss functions  described in (\ref{eq:1}). To that end, we experiment only with the cross-entropy loss using the query points ($\mathcal{L}_{{qur }}$). Subsequently, we also introduce another component of cross-entropy loss using the support set ($\mathcal{L}_{{sup }}$). The added loss term is likely to enforce the samples from the support classes to move closer around the mean embeddings, thereby helping to obtain better cluster representation which further helps in generalization. Next, we also introduce a cosine similarity-based loss term to maximize the separation between the weights matrix of the subspace, making them more discriminative.
As a result, minimizing the loss function maximizes the distances between the subspaces, which further aids in better classification. The results of these studies are presented in Table \ref{tab:Table4}.
Notice that our method outperforms the model either using $\mathcal{L}_{{qur }}$ or $\mathcal{L}_{{qur }} + \mathcal{L}_{{sup }}$ loss for nine novel classes. Hence, we conclude that each loss term in the proposed method contributes to improved performance.

\begin{table*}[t]
\centering
\scriptsize
\caption{Results of ablation studies in terms of classification accuracy with 95\% confidence interval for different combinations of the loss components. We compare our method (Proposed Loss) that comprises of ($\mathcal{L}_{\text {qur }} + \mathcal{L}_{\text {sup }} + \mathcal{L}_{\text {dis }}$) with Loss1 ($\mathcal{L}_{\text {qur }}$) and Loss2 ($\mathcal{L}_{\text {qur }} + \mathcal{L}_{\text {sup }}$) respectively.\\}
\label{tab:Table4}
\resizebox{0.8\textwidth}{!}{
\begin{tabular}{|c|c|c|c|c|}
\hline
Group  & Abnormality        & Loss1      & Loss2      & \textbf{Proposed Loss} \\ \hline
       & Fibrosis           & 45.18±1.61 & 46.74±1.91 & 42.82±1.75    \\
Group1 & Hernia             & 35.62±1.42 & 30.04±1.65 & 30.10±1.62    \\
       & Pneumonia          & 39.88±1.55 & 38.38±1.58 & 40.88±1.66    \\ \hline
       & Mass               & 34.08±1.49 & 32.42±1.45 & 35.68±1.42    \\
Group2 & Nodule             & 38.42±1.54 & 36.90±1.52 & 35.62±1.48    \\
       & Pleural Thickening & 29.88±1.38 & 30.94±1.41 & 31.08±1.41    \\ \hline
       & Cardiomegaly       & 38.18±1.51 & 37.78±1.54 & 38.20±1.54    \\
Group3 & Edema              & 65.94±1.48 & 57.66±1.54 & 65.16±1.48    \\
       & Emphysema          & 40.02±1.58 & 40.66±1.58 & 42.04±1.61    \\ \hline
       & Consolidation      & 31.84±1.56 & 35.58±1.50 & 32.84±1.58    \\
Group4 & Effusion           & 39.10±1.69 & 41.82±1.59 & 38.20±1.61    \\
       & Pneumothorax       & 45.62±1.91 & 42.72±1.62 & 49.98±1.74    \\ \hline
       & Atelectasis        & 29.60±1.38 & 30.92±1.45 & 27.98±1.44    \\
Group5 & Infiltration       & 35.08±1.49 & 35.74±1.49 & 35.94±1.50    \\
       & No   Finding       & 50.70±1.62 & 46.34±1.73 & 52.84±1.71    \\ \hline
\end{tabular}}
\end{table*}

\subsection{Role of the Random Subspace}

\begin{figure}[]
\includegraphics[width = \textwidth]{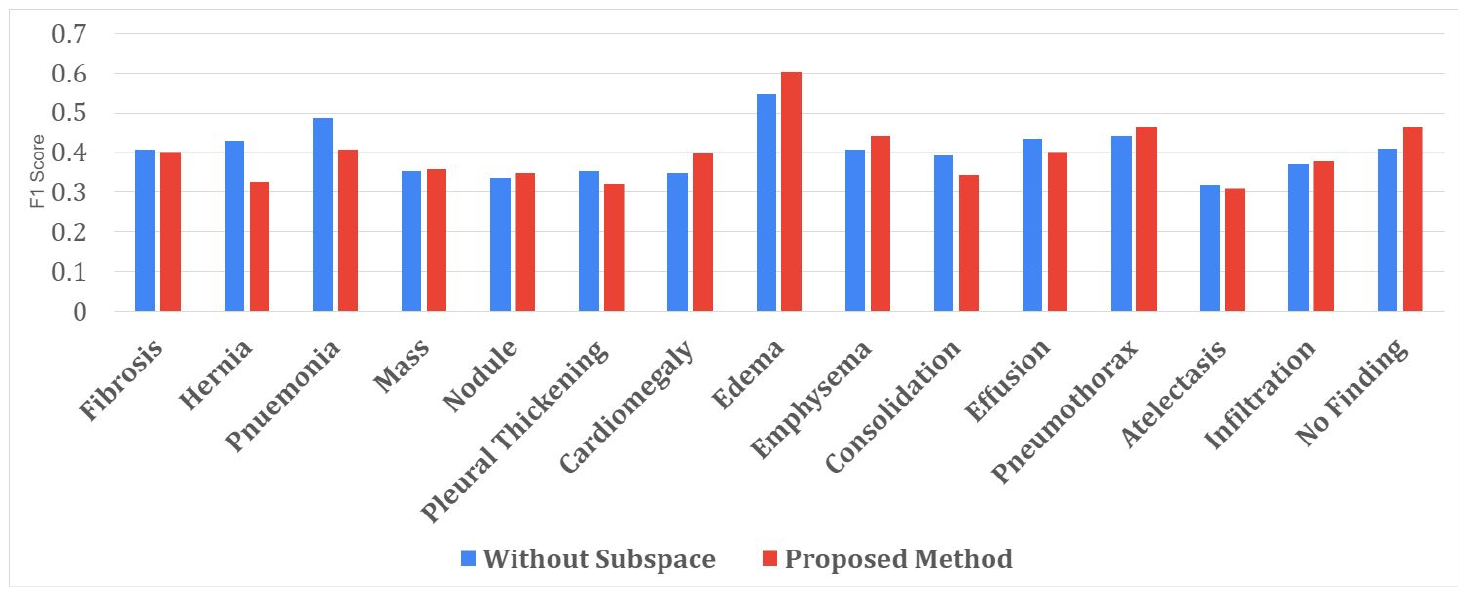} 
\caption{Comparative performance of the proposed method using random subspace (Proposed) and single feature subspace instead of random subspace (Without subspace) in terms of F1 score.}
\label{figure4}
\end{figure}

Besides selecting the appropriate number of random subspaces, we also intend to evaluate the significance of random subspaces in obtaining better performance than other methods. Accordingly, we ran the experiment with the model without subspaces and compared it with our method in terms of F1-score. From the experiment results in Fig. \ref{figure4}, it is evident that the proposed method outperforms the model with no subspace for eight novel classes while obtaining comparable results for the rest of the classes, thereby helping obtain better results.
This can be explained by the fact that the feature space contains both important and redundant features. The presence of unimportant features reduces the classifier's performance. Thus, by employing an ensemble of the subspace, we intend to investigate various combinations of features that result in improved classification accuracy.

\subsection{Choice of The Feature Extractor}

We investigate the impact of feature extractor choice by comparing our EfficientNet-B7 feature extractor to DenseNet-121 \cite{huang2017densely}. Table \ref{tab:Table5} reveals that using EfficientNet-B7 outperforms DenseNet-121 in terms of mean average accuracy for eight novel classes, while having similar performance for the others. This improvement is due to the uniform scaling of network parameters, such as depth, width, and resolution, using a compound coefficient \cite{tan2019efficientnet}. Hence, our findings suggest that EfficientNet-B7 as the feature extractor leads to better generalization for novel classes than the DenseNet-121 backbone.

\begin{table*}[h]
\centering
\caption{Accuracy of 3-way 5-shot classification with 95\% confidence interval for our proposed method (Proposed) with EfficientNet-B7 backbone alongside with DenseNet-121 feature extractor for subspace 30.}
\label{tab:my-table}
\begin{tabular}{|c|c|c|c|}
\hline
Group  & Abnormality        & DenseNet-121 & \textbf{Proposed}   \\ \hline
       & Fibrosis           & 39.78±1.69   & 42.82±1.75 \\
Group1 & Hernia             & 38.00±1.71   & 30.10±1.62 \\
       & Pneumonia          & 43.00±1.60   & 40.88±1.66 \\ \hline
       & Mass               & 33.48±1.42   & 35.68±1.42 \\
Group2 & Nodule             & 34.90±1.41   & 35.62±1.48 \\
       & Pleural Thickening & 36.24±1.46   & 31.08±1.41 \\ \hline
       & Cardiomegaly       & 37.68±1.52   & 38.20±1.54 \\
Group3 & Edema              & 59.56±1.60   & 65.16±1.48 \\
       & Emphysema          & 36.80±1.49   & 42.04±1.61 \\ \hline
       & Consolidation      & 38.56±1.52   & 32.84±1.58 \\
Group4 & Effusion           & 42.62±1.58   & 38.20±1.61 \\
       & Pneumothorax       & 37.22±1.45   & 49.98±1.74 \\ \hline
       & Atelectasis        & 29.74±1.43   & 27.98±1.44 \\
Group5 & Infiltration       & 38.54±1.58   & 35.94±1.50 \\
       & No Finding         & 43.82±1.63   & 52.84±1.71 \\ \hline
\end{tabular}
\label{tab:Table5}
\end{table*}

\section{On the Computational Complexity}
\label{sec:appendix_d}
We evaluate the time complexity of the proposed method which utilizes an ensemble of random subspaces for subspace decomposition in comparison to t-SVD. The time complexity of the t-SVD technique is $\mathcal{O}\left(2 a b^2+b^3+b+a b\right)$ \cite{li2019tutorial}, where $a$ is the number of data points and $b$ is the dimension of the feature vectors. With the assumption that the dimension of the feature vectors are constant, the time complexity can be simplified to $\mathcal{O}(a)$. On the other hand, for our method, the subspaces are generated randomly. As a result, the subspace generation process in the proposed method is independent of the number of data points. Therefore, w.r.t. the number of data points,  the time complexity of the proposed method is $\mathcal{O}( 1)$. Thus, we find that in terms of time complexity, the subspace decomposition in the proposed method has a better performance compared to the t-SVD. 

\end{document}